\documentclass{article}
\usepackage{spconf,amsmath,graphicx}
\usepackage{cite}
\usepackage{amsmath,amssymb,amsfonts}
\usepackage{graphicx}
\usepackage{textcomp}
\usepackage{bm}
\usepackage{multirow}
\usepackage{xcolor}
\usepackage{subfigure}
\usepackage{algorithm}
\usepackage{algorithmicx}
\usepackage{algpseudocode}
\usepackage{amsmath}
\usepackage{graphics}
\usepackage{epsfig}
\usepackage{url}
\usepackage{makecell}
\usepackage{xspace}

\newcommand{\vct}[1]{\boldsymbol{#1}} 
\newcommand{\mat}[1]{\boldsymbol{#1}} 

\newcommand{\vs}{\emph{vs.}\xspace}

\DeclareMathOperator*{\argmin}{arg\,min}
\newcommand{\T}{^{\textrm T}} 

\title{Meta Ordinal Weighting Net for Improving Lung Nodule Classification}
\name{Yiming Lei$^1$ \quad Hongming Shan$^{2,*}$ \quad Junping Zhang$^1$  \thanks{*Corresponding Author.}
\thanks{\textcopyright 2021 IEEE.  Personal use of this material is permitted. Permission from IEEE must be obtained for all other uses, in any current or future media, including reprinting/republishing this material for advertising or promotional purposes, creating new collective works, for resale or redistribution to servers or lists, or reuse of any copyrighted component of this work in other works.}}
\address{$^1$Shanghai Key Lab of Intelligent Information Processing, School of Computer Science\\
$^2$Institute of Science and Technology for Brain-inspired Intelligence \\ Fudan University, Shanghai 200433, China
}
\begin{document}
%
\maketitle
\begin{abstract}
The progression of lung cancer implies the intrinsic ordinal relationship of lung nodules at different stages---from \emph{benign} to \emph{unsure} then to \emph{malignant}.  This problem can be solved by ordinal regression methods, which is between classification and regression due to its ordinal label.
However, existing convolutional neural network (CNN)-based ordinal regression methods only focus on modifying classification head based on a randomly sampled mini-batch of data, ignoring the ordinal relationship resided in the data itself.
In this paper, we propose a Meta Ordinal Weighting Network (MOW-Net) to explicitly align each training sample with a meta ordinal set (MOS) containing a few  samples from all classes. During the training process, the MOW-Net learns a mapping from samples in MOS to the corresponding class-specific weight. In addition, we further propose a meta cross-entropy (MCE) loss to optimize the network in a meta-learning scheme. The experimental results demonstrate that the MOW-Net achieves better accuracy than the state-of-the-art ordinal regression methods, especially for the unsure class.
\end{abstract}

\section{Introduction}
\label{sec:Introduction}

Deep convolutional neural networks (CNNs) have recently achieved impressive results in lung nodule classification based on computed tomography (CT)~\cite{2016multiview,2017tumornet,fully3d2017,lei2020shape,icassp1,icassp2,icassp3,icassp4}. However, this kind of method usually performs a binary classification (malignant \vs benign) by omitting the unsure nodules---those between benign and malignant---which is a great waste of medical data for machine learning algorithms, especially for data-hungry deep learning methods. Therefore, how to leverage unsure data to learn a robust model is crucial for lung nodule classification.

To this end, the ordinal regression has been widely explored to utilize those unsure nodules. The unsure data model (UDM)~\cite{UDM} was proposed to learn with unsure lung nodules, and it regards this classification as an ordinal regression problem, which optimizes the negative logarithm of cumulative probabilities. However, the UDM has some additional parameters that need to be carefully tuned. The neural stick-breaking (NSB) method calculates the probabilities through the $C-1$ predicted classification bounds, where $C$ is the number of classes~\cite{liu2018ordinal}. The unimodal method makes each fully-connected output follows a unimodal distribution such as Poisson or Binomial~\cite{beckham2017unimodal}. However, these methods do not guarantee the strict ordinal relationship. Recently, the convolutional ordinal regression forests (CORFs) aim at solving this problem through the combination of CNNs and random forests~\cite{zhu2020deep}, which have been shown effective for lung nodule classification in a meta learning framework~\cite{lei2020meta}. In summary, the existing methods do not explicitly leverage the ordinal relationship resided in the data itself. 

\begin{figure}
\centerline{\includegraphics[width=1.0\linewidth]{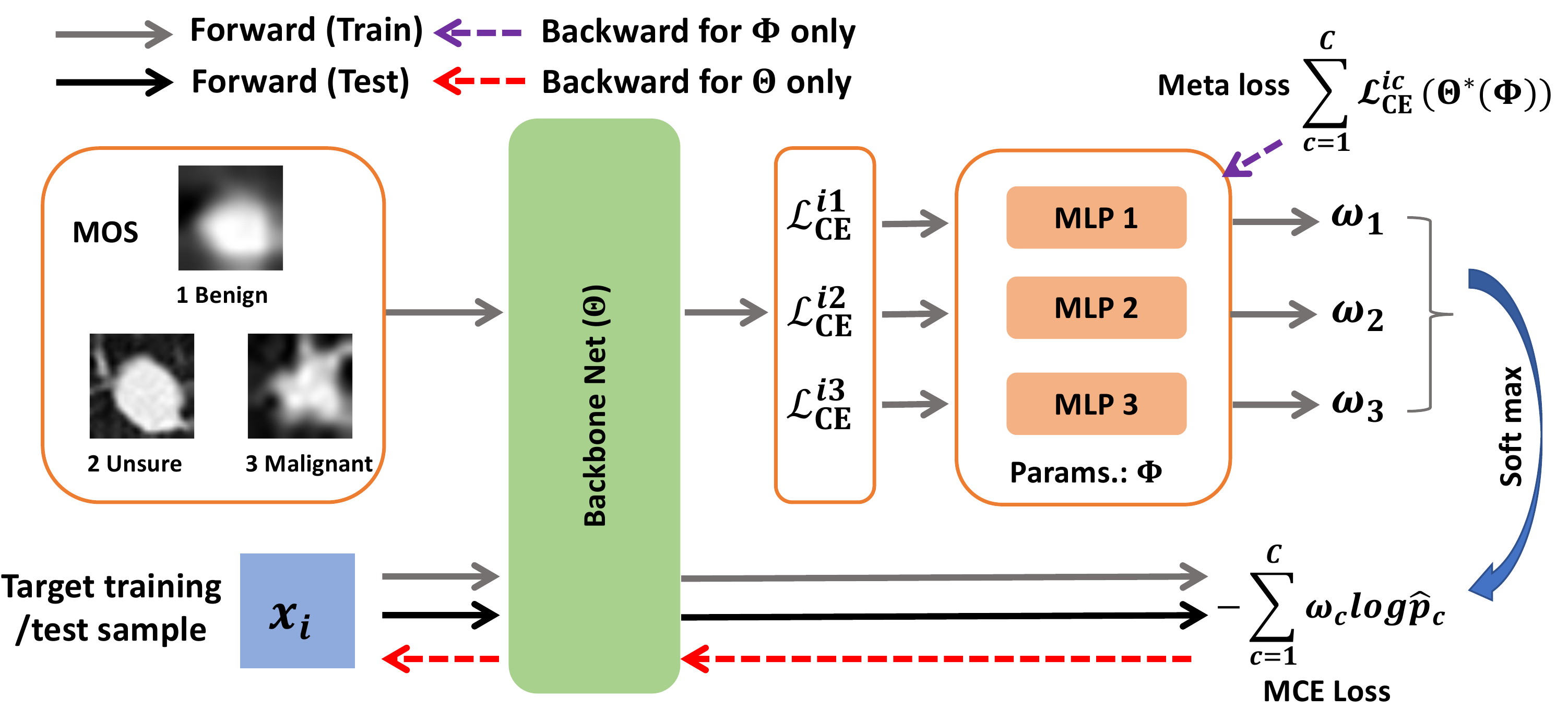}}
\caption{The illustration of the MOW-Net. Each training sample is aligned with a meta ordinal set which contains $k$ samples of each class. \textbf{Forward}: the $\mathcal{L}_{\mathrm{CE}}^{i,c}$ denotes the cross-entropy loss of class $c$ w.r.t. the $i$-th training sample, and each $\mathcal{L}_{\mathrm{CE}}^{ic}$ is mapped to the corresponding weight through the specific MLP. \textbf{Backward}: the two parts of parameters $\mat{\Theta}$ and $\mat{\Phi}$ are updated alternatively based on the proposed MCE loss and the Meta loss, respectively.}
\label{fig:framework}
\end{figure}

In this paper, we assume that the ordinal relationship resides in not only  the label but also the data itself. Put differently, the ordinal relationship of features from different classes also dominates the generalization ability of the model. Therefore, we propose a meta ordinal weighting network (MOW-Net) to learn the ordinal regression between each training sample and a set of samples of all classes simultaneously, where this set is termed as meta ordinal set (MOS). Here, the MOS implies the feature and semantic information of the dataset, and acts as the meta knowledge for the training sample. As shown in Fig.~\ref{fig:framework}, each training sample relates to an MOS, and the meta samples in the MOS are mapped to the corresponding weights representing the meta knowledge of all classes. Furthermore, we propose a novel meta cross-entropy (MCE) loss for the training of the MOW-Net; each term is weighted by the learned meta weight as shown in Fig.~\ref{fig:framework}. Different from the normal CE loss, the MCE loss indicates that the training sample is guided by the meta weights learned from the MOS. Hence, the MOW-Net is able to boost the classification performance and generalization ability with the supervision from meta ordinal knowledge. Moreover, our MOW-Net is able to reflect the difficulties of classification of all classes in the dataset, which will be analyzed in Sec.\ref{subsec:weight_analysis}.

The main contributions of this work are summarized as follows. First, we propose a meta ordinal weighting network (MOW-Net) for lung nodule classification with unsure nodules. The MOW-Net contains a backbone network for classification and a mapping branch for meta ordinal knowledge learning. Second, we propose a meta cross-entropy (MCE) loss for the training of the MOW-Net in a meta-learning scheme, which is based on a meta ordinal set (MOS) that contains a few training samples from all classes and provides the meta knowledge for the target training sample. Last, the experimental results demonstrate the significant performance improvements in contrast to the state-of-the-art ordinal regression methods. In addition,  the changes in learned meta weights reflect the difficulties of classifying each class.

\section{Methodology}
\label{sec:Methods}

This section introduces the proposed meta ordinal set (MOS), meta cross-entropy (MCE) loss, and the meta training algorithm, respectively.

\subsection{Meta Ordinal Set (MOS)}
We assume that the ordinal relationship resides in not only the label but also the data itself. Therefore, we align each target training sample with an MOS that contains $K$ samples from each class. The MOS for $i$-th training sample is formally defined as follows:
\begin{align}
\mathcal{S}_{i}^{\mathrm{meta}} = \left\{\vct{x}_{c}^{k}\right\}_{c=1,\ldots,C,\ k=1,\ldots,K},
\label{MOS}
\end{align}
where $C$ is the number of the classes.Note that the samples in the $\mathcal{S}^{\mathrm{meta}}$ are not ordered, but the samples in one class should go to the corresponding multi-layer perceptron (MLP) in Fig.~\ref{fig:framework}. Then the MPLs are able to learn the specific knowledge from each class. For a target training sample $\vct{x}_i$, $\vct{x}_{c}^{k} (c\in\{1,2,\ldots,C\})$ is randomly sampled from the training set, and  $\vct{x}_i\notin \mathcal{S}_{i}^{meta}$.

\subsection{Meta Cross-Entropy Loss}
In order to enable the MOW-Net to absorb the meta knowledge provided by the MOS, we propose an MCE loss to align the meta knowledge of each class to the corresponding entropy term:
\vspace{-2mm}
\begin{align}
\mathcal{L}_{\mathrm{MCE}} = - \sum_{c = 1}^{C} \omega_{c} \log  \hat{p}_{c},
\label{mce}
\end{align}
where $\hat{p}_{c}$ and $\omega_{c}$ are the prediction and the learned meta weight of the $c$-th class, respectively. Note that the MCE loss implies no ordinal regression tricks such as cumulative probabilities, it only holds the correlation between the meta data and the predictions. Compared with the conventional CE loss, the MCE loss enables the training samples to be supervised by the corresponding meta data, hence, the learning of the MOW-Net takes into account the meta ordinal knowledge resided in the data itself.

\subsection{Training Algorithm}

The MOW-Net is trained in the meta-learning scheme, which requires the second derivatives~\cite{shu2019meta,liu2019self,vuorio2019multimodal,maml,jamal2019task}, and two parts of the parameters, $\mat{\Theta}$ and $\mat{\Phi}$, to be updated alternatively. We highlight that the MOW-net offers a novel way of utilizing  ordinal relationship encapsulated within the data itself; however, the model is still the same as the one trained with CE loss. 
What's more, our MOW-Net does not modify the classification head and can be adapted to various backbones. 

Although the whole MOW-Net contains the two parts of parameters, $\mat{\Theta}$ and $\mat{\Phi}$, the trained model discards the MLPs ($\mat{\Phi}$) at the inference stage. In other words, the MLPs are only involved in the meta training phase to produce the class specific knowledge. Then the optimal $\mat{\Theta}$ is learned by minimizing the following objective function:
\begin{align}
\mat{\Theta}^{*}(\mat{\Phi}) &= \argmin_{\mat{\Theta}} \mathcal{L}_{\mathrm{MCE}}(\mat{\Theta};\mat{\Phi}) = \frac{1}{N} \sum_{i=1}^{N} \mathcal{L}_{\mathrm{MCE}}^{i} \label{obj:theta1}\\
						   &= - \frac{1}{N}\sum_{i=1}^{N} \sum_{c=1}^{C} \underbrace{V_{c}(\mathcal{L}_{\mathrm{CE}}^{i,c}(\mat{\Theta}); \mat{\Phi})}_{\omega_c} \cdot \log{\hat{p}}_{i,c}(\mat{\Theta}),
\notag
\end{align}
where $\mathcal{L}_{\mathrm{CE}}^{i,c}$ denotes the conventional CE loss of the $c$-th meta data with respect to the $i$-th training sample, and $V_{c}$ is the $c$-th MLP with  $\mathcal{L}_{\mathrm{CE}}^{i,c}$ being the input (Fig. \ref{fig:framework}).

Following~\cite{shu2019meta},  $\mat{\Phi}$ can be updated through the following objective function:
\begin{align}
\mat{\Phi}^{*} = \argmin_{\mat{\Phi}} \frac{1}{M}\sum_{j=1}^{M}\sum_{c=1}^{C} \mathcal{L}_{\mathrm{CE}}^{j,c}(\mat{\Theta}^{*}(\mat{\Phi})),
\label{obj:phi}
\end{align}
where $M=C\times K$ is the size of the MOS. Then we update these two parts of parameters alternatively using meta-learning~\cite{shu2019meta,liu2019self}. First, we calculate the derivative of $\mat{\Theta}$ through Eq.~\eqref{obj:theta1}:

\begin{align}
\widehat{\mat{\Theta}}^{(t)} = \mat{\Theta}^{(t)} - \frac{\alpha}{N}\sum_{i=1}^{N} \sum_{c=1}^{C} \underbrace{V_{c}(\mathcal{L}_{\mathrm{CE}}^{i,c}(\mat{\Theta}); \mat{\Phi})}_{\omega_c} \nabla_{\mat{\Theta}} \log\hat{p}_{i,c}(\mat{\Theta}) |_{\mat{\Theta}^{(t)}}. \label{eq:theta_hat}
\end{align}

Next,  $\mat{\Phi}^{(t+1)}$ can be updated as follows:
\begin{align}
\mat{\Phi}^{(t+1)} = \mat{\Phi}^{(t)} - \beta \frac{1}{M}\sum_{j=1}^{M}\sum_{c=1}^{C} \nabla_{\mat{\Phi}} \mathcal{L}_{\mathrm{CE}}^{j,c}(\widehat{\mat{\Theta}}(\mat{\Phi})) |_{\mat{\Phi}^{(t)}}. \label{eq:update_phi}
\end{align}

Last, $\mat{\Theta}^{(t+1)}$ is updated based on the $\mat{\Phi}^{(t+1)}$:
\begin{align}
\mat{\Theta}^{(t+1)} = \mat{\Theta}^{(t)} - \alpha \frac{1}{N}\sum_{i=1}^{N} \nabla_{\mat{\Theta}} \mathcal{L}_{\mathrm{MCE}}^{i}(\mat{\Theta}; \mat{\Phi}^{(t+1)}) |_{\mat{\Theta}^{(t)}} \label{eq:update_theta}.
\end{align}

In the above equations, the superscript $(t)$ represents the $t$-th iteration during training, $\alpha$ and $\beta$ are the learning rates for $\mat{\Theta}$ and $\mat{\Phi}$, respectively. The training algorithm is detailed in Algorithm~\ref{alg:train}. Note that after updating $\mat{\Phi}^{(t+1)}$ (line 8 in the Algorithm~\ref{alg:train}), the MOW-Net obtains the meta knowledge through taking the normal CE loss of the training sample as the input to all MLPs. Here, we can regard the updated MPLs as the prior knowledge for each ordinal class.

To further analyze Eq.~\eqref{eq:update_phi}, we can obtain:
\begin{align}
\mat{\Phi}^{(t+1)} &= \mat{\Phi}^{(t)} + \\
                  & \frac{\alpha\beta}{N}\sum_{i=1}^{N}(\frac{1}{M}\sum_{j=1}^{M} G_{ij}) \frac{\partial{\sum_{c=1}^{C}V_{c}(\mathcal{L}_{\mathrm{CE}}^{i,c}(\mat{\Theta}^{(t)}); \mat{\Phi})}}{\partial{\mat{\Phi}}} |_{\mat{\Phi}^{(t)}}, \notag\\
G_{i,j} &= {\left[\frac{\partial{\sum_{c=1}^{C}\mathcal{L}_{\mathrm{CE}}^{j,c}(\widehat{\mat{\Theta}})}}{\partial{\widehat{\mat{\Theta}}}} |_{\widehat{\mat{\Theta}}^{(t)}}\right]}^{\T} \cdot \frac{\partial{\sum_{c=1}^{C} \log \hat{p}_{i,c}(\mat{\Theta})}}{\partial{\mat{\Theta}}} |_{\mat{\Theta}^{(t)}}. \label{eq:G}
\end{align}
We can see that Eq.~\eqref{eq:G} denotes the derivative of the entropy of the training sample (2nd term) is to approach the derivative of the MOS (1st term). This implies that the learning of the backbone network is guided by the meta knowledge. Please refer to \url{http://hmshan.io/papers/mow-net-supp.pdf} for the detailed derivation.

\begin{algorithm}[ht]
\caption{Training Algorithm of MOW-Net.}
\label{alg:train}
 \begin{algorithmic}[1]
 \Require
 Training data $\mathcal{S}^{\mathrm{tr}}$, randomly sampled MOS $\mathcal{S}^{\mathrm{meta}}$.
 \Ensure
 The learned parameter $\mat{\Theta}^{(T-1)}$.
 \State Initialize  $\vct{\Theta}^{(0)}$ and  $\vct{\Phi}^{(0)}$.
 \For{$t=0$ to $T-1$}
  \State Sample a mini-batch of $\vct{x}_{i}$ from $\mathcal{S}^{\mathrm{tr}}$, and align each $\vct{x}_{i}$ with a MOS $\mathcal{S}^{\mathrm{meta}}_{i}$.
  \State Forward: input $\vct{x}_{i}$ and $S^{\mathrm{meta}}_{i}$ with $\vct{\Theta}^{(t)}(\vct{\Phi}^{(t)})$ to obtain $\hat{p}_{i,c}$ and $\omega_{i,c}^{meta}$ respectively.
  \State Compute the first order derivative $\widehat{\vct{\Theta}}^{(t)}(\vct{\Phi}^{(t)})$ by Eq.~\eqref{eq:theta_hat}.
  \State Forward: input $\mathcal{S}^{\mathrm{meta}}_{i}$ with $\widehat{\vct{\Theta}}^{(t)}(\vct{\Phi}^{(t)})$, then obtain the meta CE losses $\mathcal{L}_{\mathrm{CE}}^{i\_meta}$.
  \State Update $\vct{\Phi}^{(t+1)}$ by Eq.~\eqref{eq:update_phi}.
  \State Forward: input $\vct{x}_{i}$ with $\vct{\Theta}^{(t)}(\vct{\Phi}^{(t+1)})$ and calculate the $\mathcal{L}_{\mathrm{CE}}^{i\_tr}$; then input $\mathcal{L}_{\mathrm{CE}}^{i\_tr}$ into all the MLPs to obtain the $\omega_{c}^{tr}$.
  \State Update $\vct{\Theta}^{(t+1)}$ by Eq.~\eqref{eq:update_theta}.
 \EndFor
 \end{algorithmic}
\end{algorithm}

\section{Experiments}
\label{sec:Experiments}
In this section, we report the classification performance of our MOW-Net on the dataset LIDC-IDRI~\cite{LIDC}.

\subsection{Dataset}
\label{subsec:data}
LIDC-IDRI is a publicly available dataset for low dose CT-based lung nodule analysis, which includes 1,010 patients. Each nodule was rated on a scale of 1 to 5 by four thoracic radiologists, indicating an increased probability of malignancy. In this paper, the ROI of each nodule was cropped at its annotated center, with a square shape of a doubled equivalent diameter. An averaged score of a nodule was used as ground-truth for the model training. All volumes were resampled to have 1mm spacing (original spacing ranged from $0.6$mm to $3.0$mm) in each dimension, and the cropped ROIs are of the size $32 \times 32 \times 32$. The averaged scores range from 1 to 5, and in our experiments, we regard a nodule with a score between 2.5 and 3.5 as the unsure nodule; benign and malignant nodules are those with scores lower than 2.5 and higher than 3.5, respectively~\cite{UDM}.

\subsection{Implementation Details}
We used the VGG-16 as the backbone network~\cite{vgg}, and made the following changes: 1) the input channel is 32 following~\cite{lei2020shape}; 2) we only keep the first seven convolutional layers due to a small size of the input, each followed by the batch normalization (BN) and ReLU; and 3) the final classifier is a two-layer perceptron that has 4096 neurons in hidden layer. We use 80\% of data for training and the remaining data for testing.

The hyperparameters for all experiments are set as follows: the learning rate is 0.0001 and decayed by 0.1 for every 80 epochs; the mini-batch size is 16; weight decay for Adam optimizer is 0.0001~\cite{kingma2014adam}. The symbols P, R, and F1 in our results stand for precision, recall, and F1 score, respectively~\cite{UDM}.

\begin{table*}[htb]
\caption{Results of classification on LIDC-IDRI dataset. Following~~\cite{UDM}, the values with underlines indicate the best results while less important in the clinical diagnosis.}
\centering
\begin{tabular}{|c|c|*{10}{c|}}
\hline
\multirow{2}{*}{Method} & \multirow{2}{*}{Accuracy} & \multicolumn{3}{c|}{Benign} & \multicolumn{3}{c|}{Malignant} & \multicolumn{3}{c|}{Unsure} \\ \cline{3-11}
 &  & P & R & F1 & P & R & F1 & P & R & F1 \\
\hline
CE Loss & 0.517 & 0.538 & 0.668 & 0.596 & 0.562 & 0.495 & 0.526 & 0.456 & 0.360 & 0.402 \\
Poisson~\cite{beckham2017unimodal} & 0.542 & 0.548 & \underline{0.794} & 0.648 & 0.568 & 0.624 & 0.594 & 0.489 & 0.220 & 0.303 \\
NSB~\cite{liu2018ordinal} & 0.553 & 0.565 & 0.641 & 0.601 & 0.566 & 0.594 & 0.580 & 0.527 & 0.435 & 0.476 \\
UDM~\cite{UDM} & 0.548 & 0.541 & 0.767 & 0.635 & \underline{0.712} & 0.515 & 0.598 & 0.474 & 0.320 & 0.382 \\
CORF~\cite{zhu2020deep} & 0.559 & 0.590 & 0.627 & 0.608 & 0.704 & 0.495 & 0.581 & 0.476 & 0.515 & 0.495 \\
\hline
\textbf{MOW-Net ($k=1$){  }} & 0.629 & 0.752 & 0.489 & 0.592 & 0.558 & \textbf{0.851} & 0.675 & 0.600 & 0.675 & 0.635 \\
\textbf{MOW-Net ($k=5$){  }} & 0.672 & 0.764 & 0.596 & 0.670 & 0.600 & 0.802 & 0.686 & \underline{0.642} & 0.690 & 0.665 \\
\textbf{MOW-Net ($k=10$)} & \textbf{0.687} & \textbf{0.768} & 0.623 & \textbf{0.688} & 0.668 & 0.705 & \textbf{0.686} & 0.606 & \textbf{0.792} & \textbf{0.687} \\
\hline
\end{tabular}
\label{tab:results}
\end{table*}

\begin{figure*}[htb]
\centerline{\includegraphics[width=1.0\linewidth]{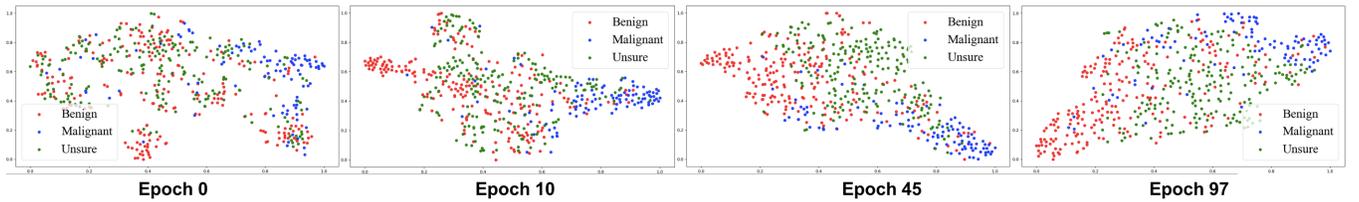}}
\vspace{-3mm}
\caption{The visualization results on the testing set using $t$-SNE.}
\label{fig:tsne}
\end{figure*}

\begin{figure}
\centerline{\includegraphics[width=1.0\linewidth]{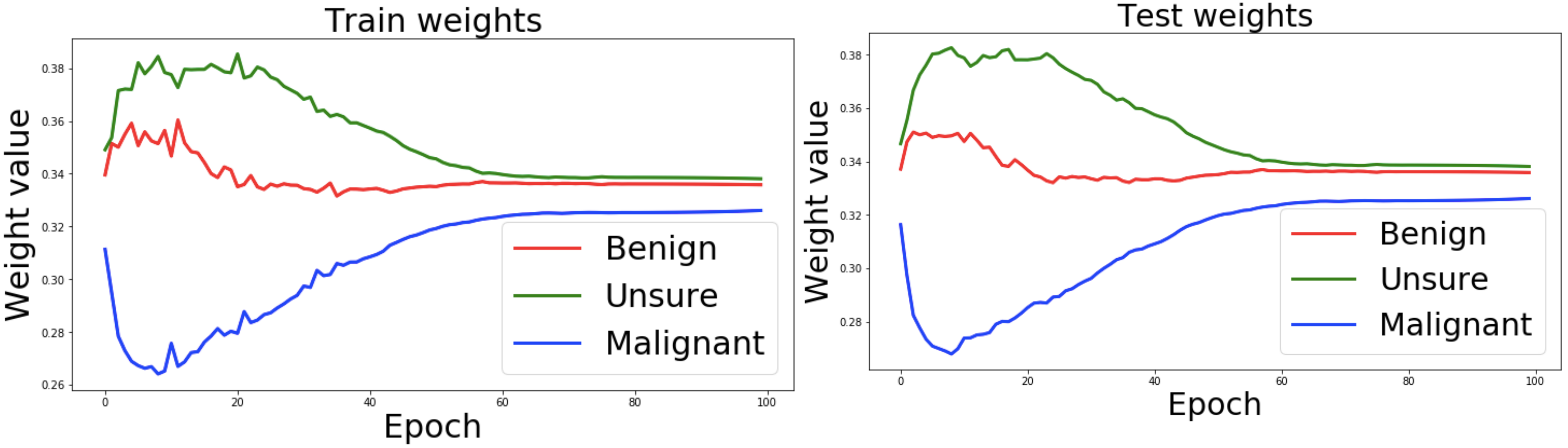}}
\vspace{-3mm}
\caption{The variations of the learned weights for all classes.}
\label{fig:weight}
\end{figure}

\subsection{Classification Performance}
In our experiments, we mainly focus on the precision of benign class, recall of malignant and unsure classes~\cite{UDM}. We  compared our MOW-Net with the state-of-the-art ordinal regression methods and the normal CE loss. In Table~\ref{tab:results}, we can see that the MOW-Net achieves the best accuracy by a large margin against other methods. Specifically, the MOW-Net significantly improves the recall of the unsure class by 0.28 over the previous best result. This is significant for the clinical diagnosis since a higher recall of the unsure class can encourage more follow-ups and reduce the probabilities of the nodules that are misdiagnosed as malignant or benign. In addition, the precision of benign and the recall of the malignant get a great improvement.

\subsection{Analysis on Learned Weights}
\label{subsec:weight_analysis}
In order to further understand the weighting scheme of the MOW-Net, we plot the variations of the weights in $\mathcal{L}_{\mathrm{MCE}}$ in Fig.~\ref{fig:weight}. At the beginning of the training, the weight for the unsure class is increasing while the weight for the malignant class is decreasing, indicating that the MOW-net focuses on classifying the unsure class from the other two classes, and the malignant class is an easy-classified class. Then, at epoch 10, the trends of these two weights become opposite.  The curve of the benign fluctuates slightly through the whole training process. At epoch 45, the weights for all the three classes begin to converge. This indicates that the model pays different attentions (weights) to different classes, and these attentions affect the update of the backbone network. At the end of the training, the model has similar sensitivities for each class.

Together with Fig.~\ref{fig:tsne}, the malignant samples are easier to be classified than the other two classes at the beginning. At epoch 10, the unsure samples are fused with other samples severely so that it has the highest weight. Simultaneously, the malignant class performs worse than that at the beginning. As the training continues, the weight for the malignant began to increase. At epoch 45, the malignant samples are clustered again and the unsure samples are more centralized than that of the previous epochs. At epoch 97, the model achieves the best accuracy, and it is obvious that the samples are distributed orderly, which demonstrates the effectiveness of the meta ordinal set.

\subsection{Effects on the Size of MOS}
The definition of the MOS in Eq.~\eqref{MOS} shows that the parameter $K$ determines the number of samples of each class. Here, we explore the effect of varied $K$. Table~\ref{tab:results} shows that when $K=10$, the MOW-Net obtained the best performance. The performance of $K=5$ and $K=10$ is better than that of $K=1$, which indicates that the more number in MOS, the better generalizability of the model.

\section{Conclusions}
\label{sec:Conclusions}
In this paper, we proposed an MOW-Net and the corresponding MOS to explore the ordinal relationship resided in the data itself for lung nodule classification in a meta-learning scheme. The experimental results empirically demonstrate a significant improvement compared to existing methods.
The visualization results further confirm the effectiveness of the weighting scheme and the learned ordinal relationship.


\end{document}